\documentclass[conference]{IEEEtran}
\IEEEoverridecommandlockouts
\usepackage{cite}
\usepackage{amsmath,amssymb,amsfonts}
\usepackage{algorithmic, algorithm}
\usepackage{graphicx}
\usepackage{textcomp}
\usepackage{xcolor}
\usepackage{balance}
\usepackage{color}
\usepackage{comment}
\usepackage{subcaption}
\usepackage{array}
\usepackage{enumitem}

\begin{document}

\title{Sky of Unlearning (SoUL): Rewiring Federated Machine Unlearning via Selective Pruning\\

}

\author{\IEEEauthorblockN{Md Mahabub Uz Zaman}
\IEEEauthorblockA{\textit{Department of Computer Science} \\
\textit{Texas Tech University}\\
Lubbock, Texas \\
m.zaman@ttu.edu}
\and
\IEEEauthorblockN{Xiang Sun}
\IEEEauthorblockA{\textit{Department of Electrical and Computer Engineering} \\
\textit{The University of New Mexico}\\
Albuquerque, New Mexico\\
sunxiang@unm.edu}
\and
\IEEEauthorblockN{Jingjing Yao}
\IEEEauthorblockA{\textit{Department of Computer Science} \\
\textit{Texas Tech University}\\
Lubbock, Texas \\
jingjing.yao@ttu.edu}
}

\maketitle

\begin{abstract}
The Internet of Drones (IoD), where drones collaborate in data collection and analysis, has become essential for applications such as surveillance and environmental monitoring. Federated learning (FL) enables drones to train machine learning models in a decentralized manner while preserving data privacy. However, FL in IoD networks is susceptible to attacks like data poisoning and model inversion. Federated unlearning (FU) mitigates these risks by eliminating adversarial data contributions, preventing their influence on the model. This paper proposes sky of unlearning (SoUL), a federated unlearning framework that efficiently removes the influence of unlearned data while maintaining model performance. A selective pruning algorithm is designed to identify and remove neurons influential in unlearning but minimally impact the model’s overall performance. Simulations demonstrate that SoUL outperforms existing unlearning methods, achieves accuracy comparable to full retraining, and reduces computation and communication overhead, making it a scalable and efficient solution for resource-constrained IoD networks.
\end{abstract}


\section{Introduction}
\label{intro}

Federated learning (FL) enables distributed model training across multiple devices without sharing raw data, preserving privacy and reducing communication costs \cite{10615935, 9352033}. This decentralized approach is particularly useful when data privacy is a concern or centralized data transmission is impractical. The Internet of Drones (IoD) \cite{9953950, gharibi2016internet, 9314881} serves as an ideal platform for FL, supporting large-scale, cooperative aerial sensing and real-time data collection across diverse environments. Unlike static sensor networks, drones in an IoD network are highly mobile, covering vast and remote areas while continuously generating valuable data. By leveraging FL, IoD networks enable drones to train models locally and share only model updates, ensuring both privacy and efficient collaborative intelligence.





FL in IoD networks enables collaborative model training across distributed drones but is vulnerable to security and privacy threats \cite{10420449}. Poisoning attacks, for example, can corrupt the global model by injecting manipulated data, leading to biased predictions. Similarly, membership inference attacks allow adversaries to determine whether specific data points were used in training, potentially exposing sensitive drone-collected information such as surveillance footage. Moreover, drone hijacking or node compromise \cite{10623000} can enable attackers to gain control over a drone and inject harmful updates into the system. These threats necessitate effective mechanisms for removing malicious or sensitive data, ensuring that compromised information does not persist in the FL model \cite{3679014}. 


Traditional approaches to removing specific data from a trained model often require retraining the model from scratch after excluding the requested data \cite{9519428}. However, this method is computationally expensive and impractical for IoD networks, where drones operate under limited processing power and communication constraints. To overcome this limitation, machine unlearning has emerged as an efficient alternative, allowing the targeted removal of data influence without the need for full retraining \cite{10488864}.

In the context of FL, federated unlearning (FU) extends machine unlearning by allowing selective data removal across distributed drones while preserving the decentralized nature of the system \cite{3679014}.  
FU in IoD networks faces several challenges. The major challenge is communication efficiency, as frequent large-scale updates between drones and the central server significantly increase bandwidth consumption, which is particularly problematic in resource-constrained environments \cite{gad2023}. Another challenge is maintaining overall model performance while removing specific data contributions, as naively eliminating updates can disrupt learned representations, shift decision boundaries, and degrade model accuracy \cite{10488864}. Addressing these challenges requires an approach that minimizes communication overhead while preserving model learning efficiency. 

To overcome these challenges, we propose a selective pruning algorithm to enhance the efficiency of our sky of unlearning (SoUL) framework while preserving model accuracy. The basic idea of selective pruning is to identify and remove only the neurons most influenced by the unlearning data while retaining those critical for learning. By precisely targeting these neurons, our approach minimizes unnecessary modifications, significantly reducing computational costs and communication overhead. The major contributions of this paper are summarized as follows.

\begin{itemize}[leftmargin=*]
	\item We propose SoUL, a FU framework for IoD networks, enabling the efficient elimination of the influence of unlearned data while preserving model performance in a decentralized learning environment.
    \item We design a selective pruning algorithm that enhances computational and communication efficiency by identifying and removing neurons that are significantly impacted by unlearning while retaining those crucial for learning.
    \item We evaluate SoUL through extensive experiments, demonstrating its accuracy and time efficiency by comparing it against existing benchmarks.
\end{itemize}

The remainder of this paper is organized as follows. Section \ref{sec:rel} surveys the existing literature. Section \ref{sec:sys} presents our proposed SoUL framework. Section \ref{sec:alg} elaborates on our designed selective pruning algorithm. Section \ref{sec:sim} shows the performance of SoUL by simulations. Finally, Section \ref{sec:con} concludes the paper.

\section{Related works}
\label{sec:rel}


FL in IoD networks has been investigated in multiple research. 
Imtiaz \textit{et al.} \cite{imteaj2021survey} conducted a comprehensive survey on federated learning (FL) for resource-constrained IoT devices, including IoD networks. Yao \textit{et al.} \cite{yao2023} explored energy-efficient FL in IoD, focusing on optimizing resource utilization. Semih and Yao \cite{scal2024} addressed the challenge of minimizing overall energy consumption in IoD while ensuring stringent latency requirements for FL training. Moudoud \textit{et al.} \cite{moudoud2024reputation} proposed a novel framework integrating multi-agent federated learning and deep reinforcement learning to enhance IoD security against emerging threats while maintaining privacy.


Existing research in federated unlearning (FU) primarily focuses on improving computational or storage efficiency \cite{xu2024machine}. Liu \textit{et al.} \cite{liu2021federaser} introduced Federaser, which stores client-specific historical updates to facilitate efficient unlearning. Liu \textit{et al.} \cite{liu2022right} developed a rapid retraining approach to fully erase specific data samples from a trained FL model. Zhang \textit{et al.} \cite{Fedrecovery} proposed FedRecovery, which removes a client’s influence by subtracting a weighted sum of gradient residuals from the global model. Hanlin \textit{et al.} \cite{FedAU} designed FedAU, an efficient FU method that integrates a lightweight auxiliary unlearning module into the training process, leveraging a simple linear operation to enable effective unlearning.


For pruning-based FU, Wang \textit{et al.} \cite{wang2022federated} applied scrubbing on model parameters to unlearn specific categories, pruning high-scoring channels to remove targeted classes in classification tasks. Pochinkov \textit{et al.} \cite{pochinkov2024dissecting} developed a statistics-based scoring system to identify and prune influential parameters in large language models, effectively facilitating unlearning.


To the best of our knowledge, the use of FU in resource-constrained IoD networks has not yet been explored. To fill this gap, we propose SoUL, a FU framework designed for IoD networks. We introduce a selective pruning method that enhances computational and communication efficiency by removing only the neurons most influenced by unlearning while preserving those essential for learning.

\section{Framework Design}
\label{sec:sys}

\begin{figure*}
	\centering
	\includegraphics[width=0.8\linewidth]{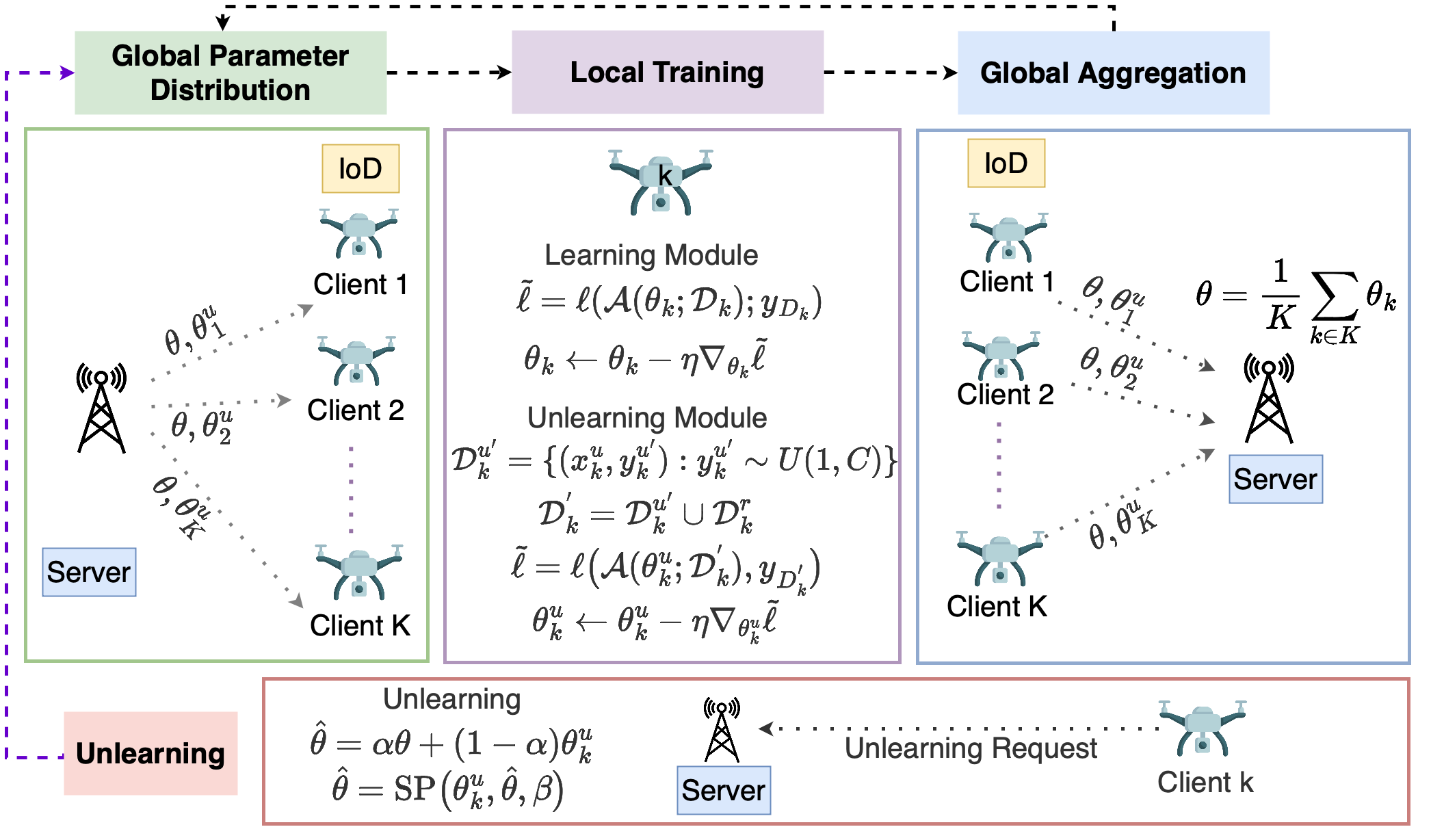}
	\caption{SoUL framework.}
    
	\label{fig:framework}
    \vspace{-10pt}
\end{figure*}

In this section, we describe our proposed SoUL framework in detail, as shown in Fig. \ref{fig:framework}. There are $K$ drones 
that act as distributed clients that locally train machine learning models using their collected data $\mathcal{D}_k$. Each drone $k$ updates its local model and periodically shares only model parameters $\theta_k$ with the central server at the ground base station (BS). The server then aggregates these models from multiple drones to refine a global model which is subsequently distributed back to the clients for further training. This iterative process continues until the model converges. 

In SoUL, drones collaboratively train a shared model $\mathcal{A}(\theta)$. The global learning objective is to minimize the loss function across all data sets, i.e., 
\begin{equation}
    \min_\theta L(\theta) = \sum_{k=1}^K \frac{|\mathcal{D}_k|}{|\mathcal{D}|} \sum_{(x_{k_i}, y_{k_i}) \in \mathcal{D}_k}  \ell(\mathcal{A}\big(\theta; x_{k_i}\big), y_{k_i}),
\end{equation}
where $\mathcal{D}_k$ is the dataset of drone $k$, $|\mathcal{D}_k|$ is the total number of samples, $|\mathcal{D}|$ is the total number of samples among all clients, $(x_{k_i}, y_{k_i}) \in \mathcal{D}_k$ is the $i$-th training data in drone $k$, $\ell(\cdot)$ is the loss function such as cross-entropy loss. 

As the system evolves, certain drones may request to unlearn a specific dataset, denoted as $D_k^u$ for drone $k$. This necessitates the modification of the federated model to exclude the influence of dataset $D_k^u$. A naive approach to accommodate this request involves retraining the local models using the remaining data $D_k^r = D_k \setminus D_k^u$, and then resubmitting these updated weights for aggregation. However, this method becomes computationally prohibitive and inefficient as the number of unlearning requests increases. To address this challenge, we develop an efficient unlearning algorithm $\mathcal{U}(\theta, D_k^u, D_k)$ to approximate the inference of retraining the model. Hence, the aim of the unlearning algorithm $\mathcal{U}(\theta, D_k^u, D_k)$ is to perform as closely as possible to the naively retrained model.

\subsection{Unlearning at Client}

To remove the influence of data $\mathcal{D}_{k}^{u}$ from $\mathcal{D}_k$ on the local model on drone $k$, an unlearning model with parameters $\theta_k^u$ is created and trained. First, drone $k$ randomly assigns labels to $\mathcal{D}_{k}^{u}$ from the class $\{1,2,...,\mathcal{C}\}$, where $\mathcal{C}$ is the total number of categories. Then, the randomly labeled data are mixed with the remaining data $\mathcal{D}_{k}^{r}$, and form a training dataset $\mathcal{D}_{k}^{'}$. Then the model is retrained using the combined dataset $\mathcal{D}_{k}^{'}$ and tries to minimize the average loss by optimizing the following objective
\begin{equation}
    \min_{\theta_k^{u}} \tilde{\ell} (\theta_k^u) = \frac{1}{|\mathcal{D}_{k}'|} \sum_{(x_{k_i}, y_{k_i}) \in \mathcal{D}_{k}'} \ell\big(\mathcal{A}({\theta_k^u}; x_{k_i}), y_{k_i}\big),
\end{equation} 
where $\mathcal{D}_k^{'}$ is the combined dataset.



\begin{algorithm}[t]
\small
\caption{SoUL}
\label{algo:unlearn-sample}
\textbf{Input:} Model $\mathcal{A}$, total client $K$, learning rate $\eta$, dataset $\mathcal{D}_{k}$ includes remaining data $\mathcal{D}_{k}^r$ and unlearning data $\mathcal{D}_{k}^u$ for unlearning client $k$, unlearning client set $\mathcal{C}_u$, pruning threshold $\beta$. 

\begin{algorithmic}[1]
\STATE Initialize the learning model $\theta$, and unlearning model $\theta_k^u$ for each client $k$;
\FOR{each global round} 
    \FOR{each drone $k$}   
        \STATE $\theta_k \gets \theta$;
        \STATE Compute the learning loss $ \ell(\mathcal{A}(\theta_k; \mathcal{D}_k); y_{\mathcal{D}_k})$;
        \STATE Update $\theta_k \leftarrow \theta_k - \eta \nabla_{\theta_k} \ell$;
    
    \IF{$k \in \mathcal{C}_u$}
        \STATE Randomly assign labels to $\mathcal{D}_k^u$ as $\mathcal{D}_k^{u'}$;
        \STATE Combine dataset $\mathcal{D}^{'}_{k} = \mathcal{D}^{u'}_{k} \cup \mathcal{D}_{k}^{r}$;
        \STATE Compute the learning loss $\tilde{\ell} = \ell(\mathcal{A}(\theta_u^k; \mathcal{D}^{'}_{k}); y_{\mathcal{D}^{'}_{k}})$;
        \STATE Update $\theta_k^u \leftarrow \theta_k^u - \eta \nabla_{\theta_k^u} \tilde{\ell}$;
    \ENDIF
    \STATE Apply $L_1$-Pruning to $\theta_k$ and $\theta_k^u$;
    \ENDFOR
    \STATE Upload the pruned $\theta_k, \theta_k^u$ to the BS;
    
    \STATE Aggregate $\theta$ as $\theta = \frac{1}{K} \sum_{k \in K}\theta_k$;
    \STATE The server implements the unlearning process:
    \\$\hat{\theta} = \alpha \theta + (1-\alpha) \theta_k^u, \ \hat{\theta} = \text{SelectivePruning}(\theta_k^u, \hat{\theta}, \beta)$;
\STATE Distribute $\theta$ to all drones.

\ENDFOR

\RETURN{$\hat{\theta}$}

\end{algorithmic}
\textbf{Output:} Unlearned parameter $\hat{\theta}$
\end{algorithm}


\subsection{Unlearning at Server}

Suppose drone $k$ makes the unlearning request to remove dataset $\mathcal{D}_{k}^{u}$ from its dataset $\mathcal{D}_{k}$. The server will use the locally trained unlearning model $\theta_k^u$ to update the global model, and then the updated global model will be distributed to all drones. 

The server updates the global model based on the following equation 
\begin{equation}
\label{eq:ul}
    \hat{\theta} = \alpha \theta + (1-\alpha) \theta_k^u,
\end{equation}
where $\alpha$ is a hyperparameter that balances between targeted unlearning accuracy and preserving the integrity of remaining data. Then, we apply a selective pruning (SP) algorithm to enhance its performance. The selective pruning algorithm selectively prunes neurons that are more significant during unlearning and less active during learning, as detailed in Section \ref{sec:alg}. If multiple drones request to unlearn, the $\theta_k^u$ is calculated as a weighted average of the unlearning parameters of $k$ drones. 



\subsection{Training Time}
\label{comm}

In each global round of SoUL, drones transmit their model parameters to the server for aggregation. Hence, the training time of each drone in one global round includes the local training time and the wireless communication time from drones to the BS. Note that the downloading time from the BS to drones is neglected in this paper because it is usually small. 

To characterize the wireless channel between drones and the BS, we adopt a widely accepted probability model that assumes that the communication channel is either line-of-sight (LoS) or Non-line-of-sight (NLoS) \cite{yao2021secure, jyaosplit23}. The probabilities of LoS and NLoS signals are given by $\Pr(\text{LoS}) = \frac{1}{1 + a e^{-b (\frac{180}{\pi}\phi - a)}}$ and $\Pr(\text{NLoS}) = 1 - \Pr(\text{LoS})$. Here, $a$ and $b$ are environmental-related constants, and $\phi$ is the elevation angle between the drone and the BS. The path losses for LoS and NLoS signals are modeled free space model. They are expressed as $PL_{LoS} = 20 \log_{10}\left(\frac{4 \pi f_c d}{c}\right) + \psi_{LoS}$ and $PL_{NLoS} = 20 \log_{10}\left(\frac{4 \pi f_c d}{c}\right) + \psi_{NLoS}$, where $\psi_{LoS}$ and $\psi_{NLoS}$ are environment-related constants, $f_c$ is the carrier frequency, $d$ is the distance between the drone and the BS, and $c$ is the speed of light. Then, the average path loss is calculated as $\overline{PL} = \Pr(\text{LoS}) PL_{\text{LoS}} + \Pr(\text{NLoS}) PL_{\text{NLoS}}$. 
The wireless channel gain between drone $k$ and the BS is given by $G_k = 10^{-\overline{PL}/10}$. According to the Shannon equation, the data transmission rate from drone $k$ to the BS can be calculated by $r_k = B \log_2\left(1 + \frac{p_k G_k}{N_0 B}\right)$, where $B$ is the bandwidth, $p_k$ is the drone $k$'s wireless transmission power, and $N_0$ is the noise power spectral. Therefore, the wireless communication time between drone $k$ and the BS is $t_k^w = \frac{s_k}{r_k}$, where $s_k$ is the size of parameters sent from drone $k$ to the BS. 

In each global round, the BS needs to receive the parameters from drones before aggregation. Hence, the global round time $T$ is determined by the training time of the slowest drone, and it can be expressed as
\begin{equation}
    T = \max_{k \in K} \left( t_k^c + t_k^w \right),
\end{equation}
where $t_k^c$ and $t_k^w$ are the local computation time and wireless communication time, respectively. 



The detailed process of SoUL is illustrated in Algorithm 1. Line 1 initializes the global model $\theta$ and unlearning models $\theta_k^u$. Lines 2-19 are the global rounds and will be repeated until convergence. In each global round, each drone updates its local learning model in Lines 4-6. If drone $k$ requests unlearning, its unlearning model is updated in Lines 7-12. Lines 15-16 aggregate the parameters. Line 17 is the unlearning process. Line 18 distributes the learning parameter to all drones.

\section{Algorithm Design}
\label{sec:alg}


In this section, we describe our selective pruning algorithm, which is designed to enhance the computation and communication efficiency of FU in IoD networks. Pruning reduces model complexity and size by removing less critical parameters of machine learning models, enhancing computational efficiency. This produces a sparse network, and fewer parameters will be transmitted to the BS, hence reducing communication overheads.

\begin{algorithm}
\caption{Selective Pruning (SP)}
\label{alg:sp-pruning}
\textbf{Input:} Learning parameters $\theta_{l}$, unlearning parameters $\theta_{ul}$, percentage $\beta$

\begin{algorithmic}[1]
    \STATE Compute weight magnitude $I_{ul}$ and $I_{l}$ for $\theta_{ul}$ and $\theta_{l}$ from their $L_1$ norm;
    
    \STATE Find the thresholds $\tau_{ul}, \tau_l$ from the $\beta$-percentile of the weight magnitude;
    
    \STATE Create masks that only allows top $\beta\%$ neurons based on their weight magnitude:
    \\\hspace*{10mm} $\mathcal{M}_{ul} \gets \mathbb{I}(I_{ul} \geq \tau_{ul}), \quad \mathcal{M}_{l} \gets \mathbb{I}(I_{l} \geq \tau_{l})$;

    \STATE Create pruning mask:
    \\ \hspace*{10mm} $\mathcal{M}_{sp} \gets
    \mathcal{M}_{ul} \setminus \mathcal{M}_{l}$;
    
    \STATE Apply mask to learning parameters:
    \\ \hspace*{10mm} $\theta_{\text{pruned}} \gets \theta_{l} \odot \neg \mathcal{M}_{sp}$;
    
    \STATE \textbf{return} pruned parameters $\theta_{\text{pruned}}$
\end{algorithmic}
\textbf{Output:} pruned parameters $\theta_{\text{pruned}}$
\end{algorithm}


The selective pruning algorithm is designed to efficiently handle unlearning requests in FU for IoD networks while preserving the model’s overall performance. When a drone requests unlearning, the server updates the global model by performing a linear aggregation of learning and unlearning weight updates. However, due to the linear nature of this operation, the decision boundary of the remaining samples shifts from its original position, which can degrade the model’s predictive performance. This shift occurs because the removal of specific data contributions alters the model’s learned feature space, affecting how the remaining data points are classified. To mitigate this issue, the selective pruning algorithm identifies and removes neurons that are more influential in the unlearning process while preserving those that are critical for the learning process.

The basic idea behind the selective pruning algorithm is based on the observation that most inputs activate only a small subset of neurons, indicating that certain neurons play a disproportionately significant role in shaping the decision boundary of the model. Removing neurons indiscriminately during unlearning can lead to unnecessary disruptions in model performance, making it crucial to selectively prune those that contribute primarily to unlearning while retaining those that maintain accuracy for the remaining data. 

The selective pruning algorithm begins by computing the weight magnitudes $I_{ul}$ for the UL parameter $\theta_{ul}$ and $I_{l}$ for the learning parameter $\theta_{l}$ using their $L_1$ norm. The $L_1$ norm is chosen because it provides a measure of the absolute importance of each weight, which allows us to identify which neurons have the strongest connection to the decision space. Next, we find thresholds $\tau_{ul}$ and $\tau_{l}$ based on the $\beta$ percentile of the weight magnitudes. The parameter $\beta$ represents the percentage of neurons we want to keep. This step allows us to identify the top $\beta\%$ most important neurons for both the learning and unlearning processes. Then, we create binary masks $\mathcal{M}_{ul}$ and $\mathcal{M}_{l}$ for the unlearning and learning parameters, respectively. These masks are created using indicator function $\mathbb{I(\cdot)}$, which returns 1 for weights above the thresholds and 0 otherwise. The masks for unlearning and learning are represented as $\mathcal{M}_{ul} = \mathbb{I}(I_{ul} \geq \tau_{ul})$, $\mathcal{M}_{l} = \mathbb{I}(I_{l} \geq \tau_{l})$, where $\mathbb{I}(\cdot)$ is the indicator function, $I_{ul}$ and $I_{l}$ the weight magnitudes, and $\tau_{ul}$ and $\tau_{l}$ are the thresholds. 
After that, the pruning mask $\mathcal{M}_{sp}$ is created by finding the set difference between  $\mathcal{M}_{ul}$ and  $\mathcal{M}_{l}$. This operation identifies neurons that are important for unlearning but not as important for learning. Finally, we apply the pruning mask to the learning parameters $\theta_l$ by element-wise matrix multiplication $\odot$ between the learning parameter $\theta_{l}$ and the negation of selective pruning mask $\mathcal{M}_{sp}$. This effectively zeros out the weights of neurons identified for pruning while keeping the weights of important neurons unchanged.


The detailed process of our proposed selective pruning algorithm is illustrated in Algorithm \ref{alg:sp-pruning}. Line 1 computes the weight magnitude $I_{ul}$ and $I_{l}$ for unlearning parameter and learning parameter, respectively. Lines 2-4 create the pruning mask considering the $\beta$-percentile threshold. Finally, Lines 5-6 apply the pruning mask to the learning parameter $\theta_{l}$.

\section{Performance Evaluation}
\label{sec:sim}
In this section, we set up simulations to evaluate the performance of our proposed framework SoUL. The simulation is conducted with a quad-core Intel Xeon Gold 6242 Processor, an NVIDIA Tesla V100 16 GB GPU, and 36 GB of memory. We compare SoUL with two benchmark algorithms, including Retrain \cite{alex2009learning} and FedAU \cite{FedAU}. Retrain trains the model with the remaining data after removing the requested data. FedAU is an FU framework that does not consider the model sparsity relationship and selective pruning. 


In the IoD network, there are 50 drones randomly distributed within a $10000 \ \textit{m} \times 10000 \ \textit{m}$ area. For wireless channels, the environmental constants are set as \textit{$a=9.6$}, \textit{$b=0.28$}, \textit{$\psi_{\text{LoS}}=1$~dB}, and \textit{$\psi_{\text{NLoS}}=20$~dB}. The carrier frequency is \textit{$f_c=2$~GHz}, the bandwidth is \textit{$B=2$~MHz}, the noise density is \textit{$N_0=-174$~dBm/Hz}, the speed of
light \textit{$c = 3 \times 10^8$ m/s} and the maximum transmit power is \textit{$p_k=3$~W}. The height of the drones \textit{$H = 100$ m}. The above parameters related to drone wireless communications are consistent with \cite{yao2021secure}. The size $s_k$ of the transmitted parameters before $L_1$-pruning is \textit{$10$~MB} while the size becomes \textit{$2.5$~MB} after $L_1$-pruning. 




We utilize the CIFAR-10 \cite{alex2009learning} dataset for training, which contains 60,000 color images, divided between 50,000 for training and 10,000 for testing. Each image in CIFAR-10 has a resolution of 32x32 pixels and is classified into one of 10 categories, including animals (e.g., cats, dogs, horses) and vehicles (e.g., airplanes, cars, ships). AlexNet \cite{krizhevsky2012imagenet} is used for classification. The key parameters of model training are listed in Table \ref{tab_param}.

\begin{table}[t]
    \centering
    \caption{Simulation Parameters}
    \label{tab_param}
    \begin{tabular}{|>{\raggedright}p{4.5cm}|p{3cm}|}
        \hline
        \textbf{Parameter} & \textbf{Values} \\ \hline
        Optimization method & SGD \\ \hline
        Learning rate & $1 \times 10^{-2}$ \\ \hline
        Weight decay & $4 \times 10^{-5}$ \\ \hline
        Batch size & 32 \\ \hline
        Local Episode & 2 \\ \hline
        Round & 200 \\ \hline
        Unlearning data ratio & [1\% - 10\%] \\ \hline
        The number of unlearning clients & [5 - 25] \\ \hline
        Coefficient, $(1-\alpha)$ & [.65 - .90] \\ \hline
        $\beta$ for Selective Pruning & 0.20 \\ \hline
    \end{tabular}
    \vspace{-5pt}
\end{table}

\begin{figure}
    \centering
    \includegraphics[width=0.78\linewidth]{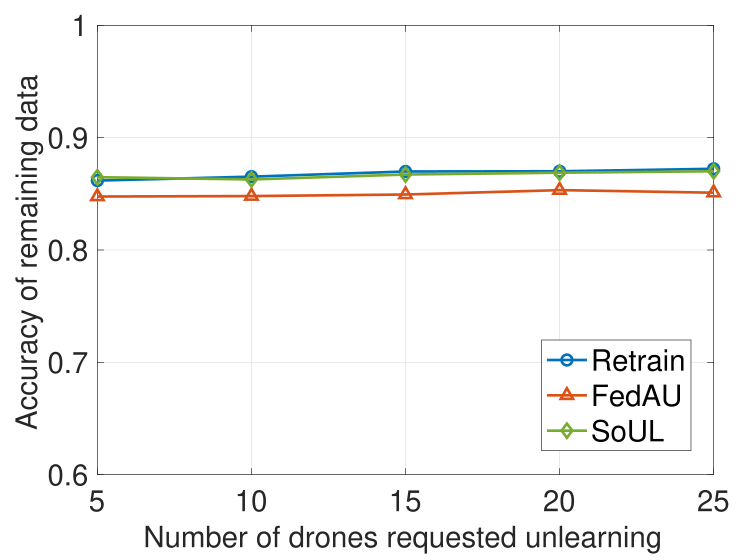}   
    \caption{Accuracy of remaining data vs number of drones requested unlearning. 
    }
    \label{fig:performance}
    \vspace{-15pt}
\end{figure}

Fig. \ref{fig:performance} illustrates the performance of accuracy with different numbers of drones requested unlearning ranging from 5 to 25. We can observe that the SoUL method achieves 87\% accuracy, outperforming FedAU and closely matching the performance of retraining. This suggests that while the unlearning requests cause a shift in the decision boundary for remaining samples, resulting in the performance decline of FedAU, our proposed SoUL mitigates this impact through its selective pruning algorithm, effectively preserving model accuracy. Moreover, the accuracy is not significantly affected by the number of drones requested unlearning. This is because the unlearning process is performed on the server through a linear operation between the learning parameter and the unlearning parameter, and so there is minimal influence by the number of clients.

\begin{figure}
    \centering
    \includegraphics[width=0.8\linewidth]{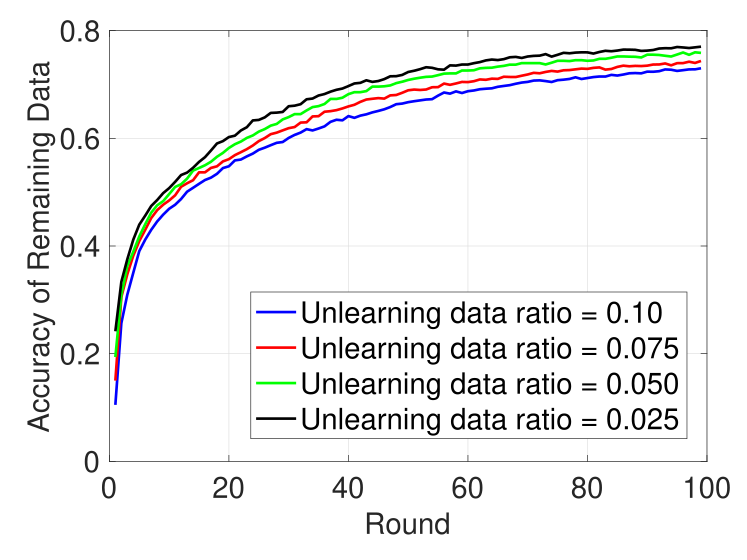}   
    \caption{Accuracy of remaining data vs round.}
    \label{fig:prop}
 \vspace{-15 pt}
\end{figure}

Fig. \ref{fig:prop} illustrates the accuracy of SoUL over time under different unlearning data ratios, ranging from 0.025 to 0.10. The UL data ratio represents the proportion of data requested for removal relative to the total dataset across all clients. We observed that the model’s accuracy declines as the unlearning data ratio increases. This trend occurs because a higher unlearning ratio results in the removal of a larger portion of training data, reducing the amount of useful information available for learning and ultimately leading to lower model performance.




\begin{figure*}
    \centering
    \begin{subfigure}[b]{0.3\linewidth}
        \includegraphics[width=\linewidth]{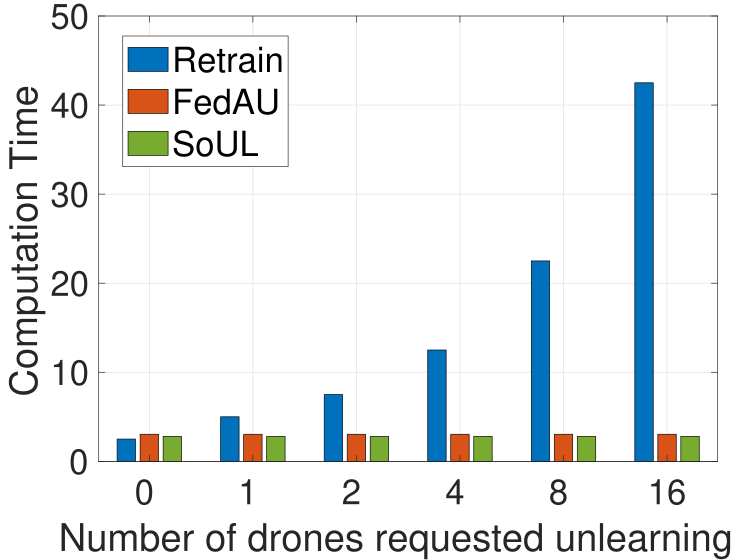}
        \caption{Computation time}
        \label{fig:comp_time}
    \end{subfigure}
    \hfill
    \begin{subfigure}[b]{0.3\linewidth}
        \includegraphics[width=\linewidth]{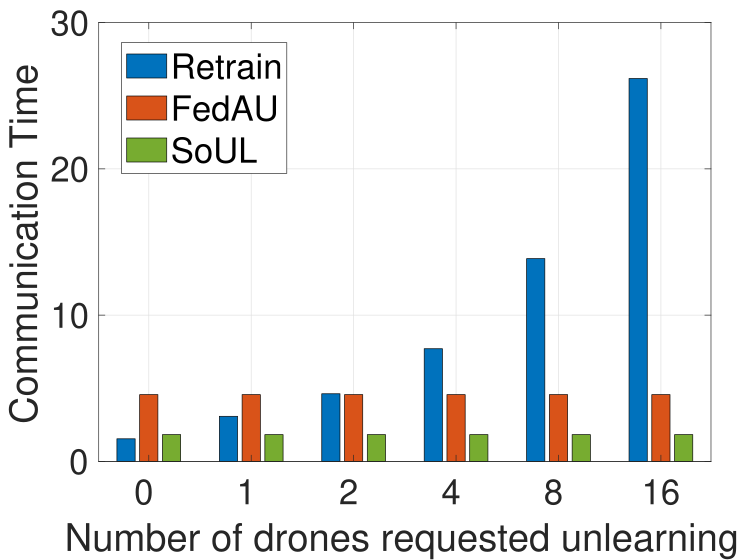}
        \caption{Communication time}
        \label{fig:comm_time}
    \end{subfigure}
    \hfill
    \begin{subfigure}[b]{0.3\linewidth}
        \includegraphics[width=\linewidth]{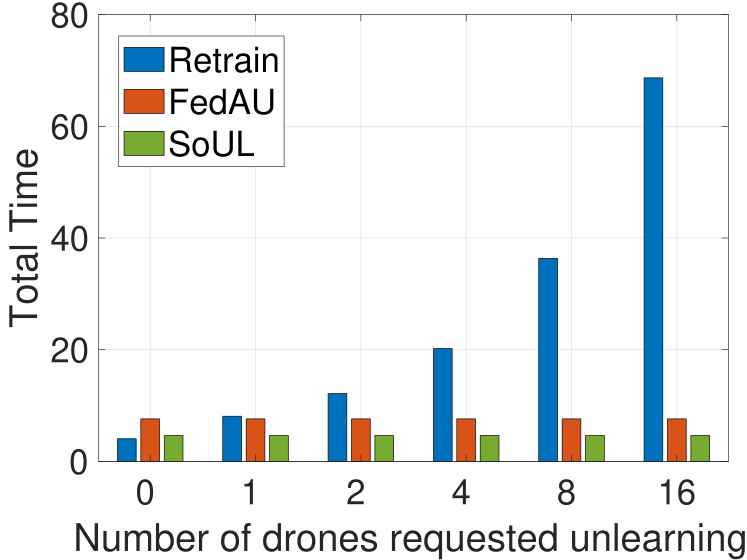}
        \caption{Total time}
        \label{fig:total_time}
    \end{subfigure}
    \caption{Time vs number of drones requested unlearning.}
    \label{fig:time_analytics2}
\end{figure*}

Fig. \ref{fig:time_analytics2} presents the computation time, communication time, and total time for three different algorithms under varying numbers of drones requesting unlearning. These times are measured per global round. The results show that Retrain experiences an exponential increase in all three metrics as the number of unlearning requests grows, making it computationally impractical for large-scale deployments. In contrast, SoUL and FedAU maintain relatively stable computation and communication times, demonstrating their efficiency. Notably, SoUL achieves lower computation and communication times than FedAU, attributed to its selective pruning algorithm, which optimizes unlearning by removing only the most relevant parameters. Overall, SoUL reduces total training time by approximately 40\% compared to FedAU, demonstrating its advantage in computation and communication efficiency.

\section{Conclusion}
\label{sec:con}

In this paper, we have proposed SoUL, a federated unlearning framework in IoD networks. We have designed a selective pruning algorithm that eliminates neurons primarily influenced by unlearning while preserving those essential for learning. Our simulation results demonstrate that the accuracy of SoUL outperforms the existing FedAU method and closely matches the accuracy of full retraining Retrain. Moreover, SoUL significantly reduces both computation and communication time, demonstrating its efficiency in unlearning while ensuring scalability in resource-constrained IoD networks. 

\bibliographystyle{IEEEtran}
\bibliography{SoUL_final}

\begin{thebibliography}{10}
\providecommand{\url}[1]{#1}
\csname url@samestyle\endcsname
\providecommand{\newblock}{\relax}
\providecommand{\bibinfo}[2]{#2}
\providecommand{\BIBentrySTDinterwordspacing}{\spaceskip=0pt\relax}
\providecommand{\BIBentryALTinterwordstretchfactor}{4}
\providecommand{\BIBentryALTinterwordspacing}{\spaceskip=\fontdimen2\font plus
\BIBentryALTinterwordstretchfactor\fontdimen3\font minus \fontdimen4\font\relax}
\providecommand{\BIBforeignlanguage}[2]{{%
\expandafter\ifx\csname l@#1\endcsname\relax
\typeout{** WARNING: IEEEtran.bst: No hyphenation pattern has been}%
\typeout{** loaded for the language `#1'. Using the pattern for}%
\typeout{** the default language instead.}%
\else
\language=\csname l@#1\endcsname
\fi
#2}}
\providecommand{\BIBdecl}{\relax}
\BIBdecl

\bibitem{10615935}
S.~Cal, X.~Sun, and J.~Yao, ``Energy-efficient federated knowledge distillation learning in internet of drones,'' in \emph{2024 IEEE International Conference on Communications Workshops (ICC Workshops)}, 2024, pp. 1256--1261.

\bibitem{9352033}
O.~A. Wahab, A.~Mourad, H.~Otrok, and T.~Taleb, ``Federated machine learning: Survey, multi-level classification, desirable criteria and future directions in communication and networking systems,'' \emph{IEEE Communications Surveys \& Tutorials}, vol.~23, no.~2, pp. 1342--1397, 2021.

\bibitem{9953950}
J.~Yao and N.~Ansari, ``{QoS}-aware machine learning task offloading and power control in internet of drones,'' \emph{IEEE Internet of Things Journal}, vol.~10, no.~7, pp. 6100--6110, 2023.

\bibitem{gharibi2016internet}
M.~Gharibi, R.~Boutaba, and S.~L. Waslander, ``Internet of drones,'' \emph{IEEE Access}, vol.~4, pp. 1148--1162, 2016.

\bibitem{9314881}
J.~Yao and N.~Ansari, ``Wireless power and energy harvesting control in {IoD} by deep reinforcement learning,'' \emph{IEEE Transactions on Green Communications and Networking}, vol.~5, no.~2, pp. 980--989, 2021.

\bibitem{10420449}
E.~Hallaji, R.~Razavi-Far, M.~Saif, B.~Wang, and Q.~Yang, ``Decentralized federated learning: A survey on security and privacy,'' \emph{IEEE Transactions on Big Data}, vol.~10, no.~2, pp. 194--213, 2024.

\bibitem{10623000}
Y.~Mekdad, A.~Acar, A.~Aris, A.~El~Fergougui, M.~Conti, R.~Lazzeretti, and S.~Uluagac, ``Exploring jamming and hijacking attacks for micro aerial drones,'' in \emph{IEEE International Conference on Communications}, 2024, pp. 1939--1944.

\bibitem{3679014}
\BIBentryALTinterwordspacing
Z.~Liu, Y.~Jiang, J.~Shen, M.~Peng, K.-Y. Lam, X.~Yuan, and X.~Liu, ``A survey on federated unlearning: Challenges, methods, and future directions,'' \emph{ACM Comput. Surv.}, vol.~57, no.~1, Oct. 2024. [Online]. Available: \url{https://doi.org/10.1145/3679014}
\BIBentrySTDinterwordspacing

\bibitem{9519428}
L.~Bourtoule, V.~Chandrasekaran, C.~A. Choquette-Choo, H.~Jia, A.~Travers, B.~Zhang, D.~Lie, and N.~Papernot, ``Machine unlearning,'' in \emph{2021 IEEE Symposium on Security and Privacy (SP)}, 2021, pp. 141--159.

\bibitem{10488864}
J.~Xu, Z.~Wu, C.~Wang, and X.~Jia, ``Machine unlearning: Solutions and challenges,'' \emph{IEEE Transactions on Emerging Topics in Computational Intelligence}, vol.~8, no.~3, pp. 2150--2168, 2024.

\bibitem{gad2023}
G.~Gad, A.~Farrag, Z.~M. Fadlullah, and M.~M. Fouda, ``Communication-efficient federated learning in drone-assisted iot networks: Path planning and enhanced knowledge distillation techniques,'' \emph{IEEE International Symposium on Personal, Indoor and Mobile Radio Communications (PIMRC)}, pp. 1--7, 2023.

\bibitem{imteaj2021survey}
A.~Imteaj, U.~Thakker, S.~Wang, J.~Li, and M.~H. Amini, ``A survey on federated learning for resource-constrained iot devices,'' \emph{IEEE Internet of Things Journal}, vol.~9, no.~1, pp. 1--24, 2022.

\bibitem{yao2023}
J.~Yao and X.~Sun, ``Energy-efficient federated learning in internet of drones networks,'' in \emph{2023 IEEE 24th International Conference on High Performance Switching and Routing (HPSR)}, 2023, pp. 185--190.

\bibitem{scal2024}
S.~Cal, X.~Sun, and J.~Yao, ``Energy-efficient federated knowledge distillation learning in internet of drones,'' pp. 1256--1261, 2024.

\bibitem{moudoud2024reputation}
H.~Moudoud, Z.~A. El~Houda, and B.~Brik, ``Reputation-aware scheduling for secure internet of drones: A federated multi-agent deep reinforcement learning approach,'' in \emph{IEEE INFOCOM 2024 - IEEE Conference on Computer Communications Workshops}, 2024, pp. 1--6.

\bibitem{xu2024machine}
M.~Xu, ``Machine unlearning: challenges in data quality and access,'' in \emph{Proceedings of the Thirty-Third International Joint Conference on Artificial Intelligence}, ser. IJCAI '24, 2024.

\bibitem{liu2021federaser}
G.~Liu, X.~Ma, Y.~Yang, C.~Wang, and J.~Liu, ``Federaser: Enabling efficient client-level data removal from federated learning models,'' in \emph{2021 IEEE/ACM 29th International Symposium on Quality of Service (IWQOS)}, 2021, pp. 1--10.

\bibitem{liu2022right}
Y.~Liu, L.~Xu, X.~Yuan, C.~Wang, and B.~Li, ``The right to be forgotten in federated learning: An efficient realization with rapid retraining,'' in \emph{IEEE INFOCOM 2022 - IEEE Conference on Computer Communications}, 2022, p. 1749–1758.

\bibitem{Fedrecovery}
L.~Zhang, T.~Zhu, H.~Zhang, P.~Xiong, and W.~Zhou, ``{FedRecovery}: Differentially private machine unlearning for federated learning frameworks,'' \emph{IEEE Transactions on Information Forensics and Security}, vol.~18, pp. 4732--4746, 2023.

\bibitem{FedAU}
H.~Gu, G.~Zhu, J.~Zhang, X.~Zhao, Y.~Han, L.~Fan, and Q.~Yang, ``Unlearning during learning: an efficient federated machine unlearning method,'' in \emph{Proceedings of the Thirty-Third International Joint Conference on Artificial Intelligence}, ser. IJCAI '24, 2024.

\bibitem{wang2022federated}
J.~Wang, S.~Guo, X.~Xie, and H.~Qi, ``Federated unlearning via class-discriminative pruning,'' in \emph{Proceedings of the ACM Web Conference 2022}, ser. WWW '22.\hskip 1em plus 0.5em minus 0.4em\relax New York, NY, USA: Association for Computing Machinery, 2022, p. 622–632.

\bibitem{pochinkov2024dissecting}
N.~Pochinkov and N.~Schoots, ``Dissecting language models: Machine unlearning via selective pruning,'' \emph{preprint arXiv:2403.01267}, 2024.

\bibitem{yao2021secure}
J.~Yao and N.~Ansari, ``Secure federated learning by power control for internet of drones,'' \emph{IEEE Transactions on Cognitive Communications and Networking}, vol.~7, no.~4, pp. 1021--1031, 2021.

\bibitem{jyaosplit23}
J.~Yao, ``Split learning for image classification in internet of drones networks,'' in \emph{2023 IEEE 24th International Conference on High Performance Switching and Routing (HPSR)}, 2023, pp. 52--55.

\bibitem{alex2009learning}
A.~Krizhevsky, G.~Hinton \emph{et~al.}, ``Learning multiple layers of features from tiny images,'' 2009.

\bibitem{krizhevsky2012imagenet}
A.~Krizhevsky, I.~Sutskever, and G.~E. Hinton, ``Imagenet classification with deep convolutional neural networks,'' \emph{Communications of the ACM}, vol.~60, no.~6, p. 84–90, May 2017.

\end{thebibliography}

\end{document}